\documentclass[letterpaper, 10 pt, conference]{ieeeconf}  

\IEEEoverridecommandlockouts                              
\usepackage{xcolor}
\usepackage{amsmath}
\usepackage{booktabs}
\usepackage{amsfonts}
\usepackage{graphicx}
\usepackage{algorithmic}
\usepackage{algorithm}
\graphicspath{ {./Assets/} }

\usepackage{hyperref}

\newcommand{\ec}[1]{{\color{black}#1}}

\title{\LARGE \bf
Learning Video-Conditioned Policies for Unseen Manipulation Tasks
}

\author{Elliot Chane-Sane$^{1}$, Cordelia Schmid$^{1}$, Ivan Laptev$^{1}$
\thanks{
\ec{$^{1}$Inria, \'{E}cole normale sup\'{e}rieure, CNRS, PSL Research University, 75005 Paris, France.
Corresponding author: elliot.chane-sane@inria.fr
\newline
Project website: \url{https://www.di.ens.fr/willow/research/vip/}
}}%
}

\begin{document}

\maketitle
\thispagestyle{empty}
\pagestyle{empty}

\begin{abstract}
The ability to specify robot commands by a non-expert user is critical for building generalist agents capable of solving a large variety of tasks.
One convenient way to specify the intended robot goal is by a video of a person demonstrating the target task.
While prior work typically aims to imitate human demonstrations performed in robot environments, here we focus on a more realistic and challenging setup with demonstrations recorded in natural and diverse human environments. 
We propose \textit{Video-conditioned Policy learning~(ViP)}, a data-driven approach that maps human demonstrations of previously unseen tasks to robot manipulation skills.
To this end, we learn our policy to generate appropriate actions given current scene observations and a video of the target task. 
To encourage generalization to new tasks, we avoid particular tasks during training and learn our policy from unlabelled robot trajectories and corresponding robot videos. Both robot and human videos in our framework are represented by video embeddings pre-trained for human action recognition.
At test time we first translate human videos to robot videos in the common video embedding space, and then use resulting embeddings to condition our policies.
Notably, our approach enables robot control by human demonstrations in a \textit{zero-shot manner}, i.e., without using robot trajectories paired with human instructions during training.
We validate our approach on a set of challenging multi-task robot manipulation environments and outperform state of the art.
Our method also demonstrates excellent performance in a new challenging zero-shot setup where no paired data is used during training.
\end{abstract}

\section{Introduction}\label{sec:introduction}
Significant \ec{progress has }
been made in recent years towards learning a generalist robot agent capable of accomplishing a wide array of skills across many environments~\cite{ahn2022can,jang2022bc,lu2022aw,chebotar2021actionable}.
Central to this challenge is the ability to effectively specify tasks and rewards to the robot system in a user-friendly manner.
In reinforcement learning, a task is commonly defined through a reward function~\cite{sutton2018reinforcement}.
However, designing good reward functions for each task is often challenging and restricting policy learning to a fixed set of tasks hinders generalization to new tasks.
Goal-conditioned imitation and reinforcement learning~\cite{ghosh2019learning,lynch2020learning,eysenbach2020c,nair2018visual,gupta2019relay,pong2018temporal} can learn agents capable of performing a wide diversity of tasks with less supervision. 
But providing the right goal to define the task requires expert operators to come up with a suitable robot observation of the desired configuration.
Other works have shown that we can learn to command generalist robots through language instructions~\cite{jang2022bc,lynch2020language,nair2022learning, mees2022matters} and human videos~\cite{jang2022bc, chen2021learning}, which are easy to provide for non-expert operators and can generalize to unseen inputs, including behaviors beyond goal-reaching skills. 
Furthermore, being able to specify robot skills through language or video commands unlocks solving more complex long-horizon tasks by chaining human instructions~\cite{ahn2022can, mees2022calvin}.
Nonetheless, these methods rely on annotated demonstrations for a large set of robot skills, which is often tedious to provide, especially since task annotation must be repeated for each new robot environment.

\begin{figure}[t]
  \centering
  \includegraphics[width=1.0\linewidth]{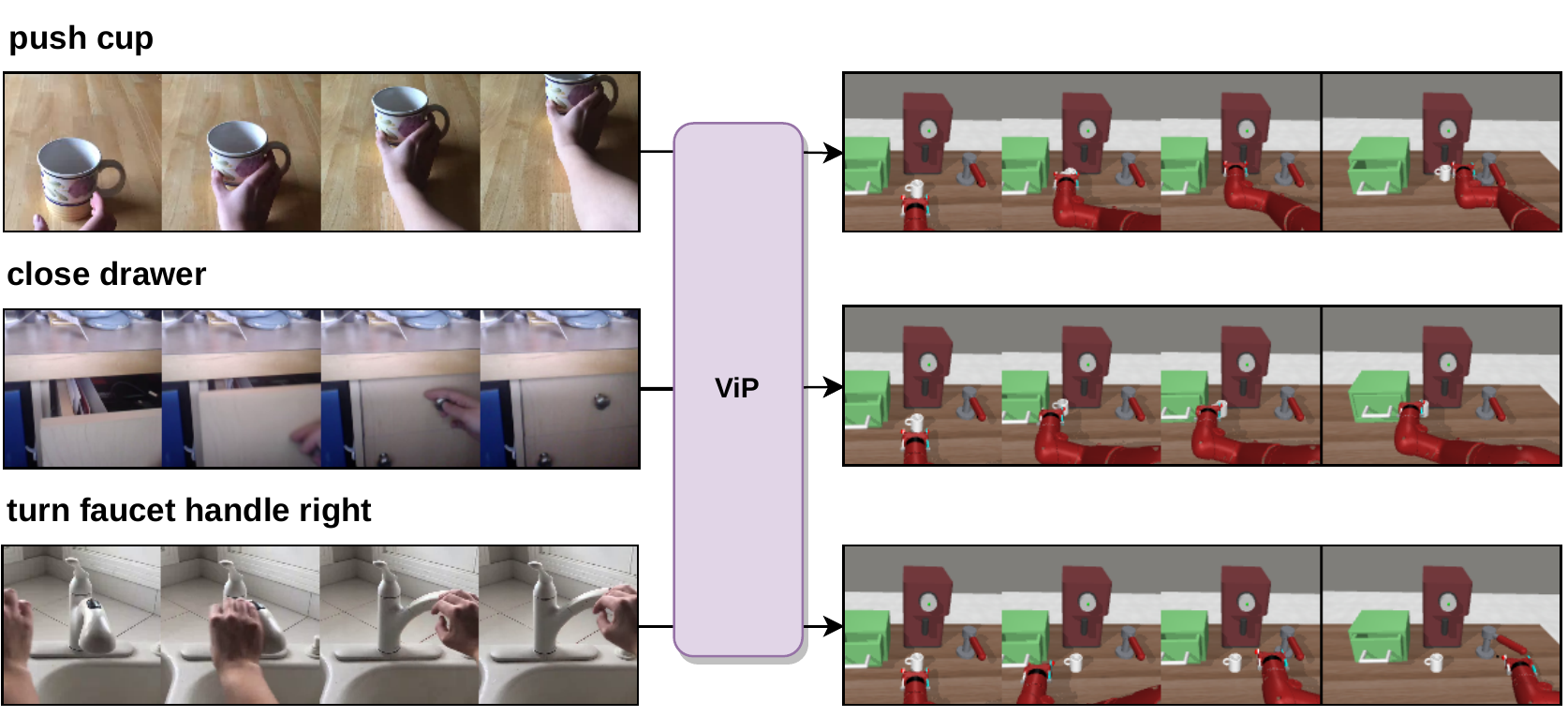}
  \vspace{-0.7cm} 
  \caption{
  Given a human video instruction in a non-robotic scene, our video-conditioned policy ViP controls the robot to perform a similar task zero-shot, i.e. the agent never observes robot data paired with human instructions during training and figures out which manipulation skill it is expected to perform in its environment at test time. We illustrate three examples of different tasks demonstrated by people and the corresponding roll-outs generated by our method for the TableTop robotics environment. \vspace{-.5cm}  
  }
  \label{fig:Teaser}
   \vspace{-0.1cm} 
\end{figure}

In this work, we propose \textit{\textbf{Vi}deo-conditioned \textbf{P}olicy learning} (ViP), a method that learns to perform manipulation skills given a human video of the desired task in vision-based multi-task robotic environments (Figure \ref{fig:Teaser}).
We demonstrate that, due to the similarity between robot manipulation and videos of humans performing manipulation skills, we can leverage existing large datasets of annotated human videos, such as the Something-Something-v2 dataset~\cite{goyal2017something} (SSv2), to learn to map human videos to robot behaviors \textit{in a zero-shot manner}, without training on paired data between human instructions and robot demonstrations.
For instance, the robot may interact in an environment that includes a drawer, yet has received no supervision on what it is expected to do when commanded with a video instruction of a human closing a possibly different drawer.
We do so by learning a video encoder using Supervised Contrastive Learning~\cite{khosla2020supervised}, where embeddings of videos of the same task are closer together in cosine distance, and show that video models trained on such large datasets of annotated human videos, which can be easily collected from the internet~\cite{goyal2017something,miech2019howto100m,grauman2022ego4d,damen2018scaling}, generalize to the robot video domain.

In addition, we show that video encoders trained for human video action recognition readily provide relevant task embeddings for multi-task policy learning.
Following recent trends in data-driven robotics, our approach learns from large collections of offline robot experience that can be collected in different ways, e.g. by expert demonstrations given by motion planners, teleoperated play data, random data generation processes.
Given an offline dataset of robot demonstrations, we learn a video conditioned policy to regress the action conditioned on both the robot state and a video embedding of the full trajectory the state-action pair belongs to.
We also keep each embedding in a library of robot embeddings.
At inference, we perform nearest neighbors regression on this library using the cosine distance to the embedding of the human instruction to generate an appropriate robot embedding that is both (a)~relevant to the instruction and (b)~is executable by the policy. 
We can then execute the human video instruction by decoding the selected embedding into a robot trajectory using the learned policy.

Overall, our approach demonstrates that large collections of human videos can enable less supervision in data-driven robotics.
Our contributions can be summarizes as follows:
\begin{itemize}
\item we designed a method to train a policy conditioned on robot video embeddings given by a video encoder pretrained on a large dataset of annotated human videos;
\item our method can map human video instructions to robot manipulation skills without supervision from paired data to bridge the gap between the human and robot domain;
\item our approach outperforms prior works on a set of multi-task robotic environments.
\end{itemize}

\section{Related Work}\label{sec:related_work}

Different methods have been explored in recent years towards robot learning with human videos.
Many prior works hence consider the problem of learning to follow human demonstrations using different techniques such as pose or keypoints estimation \cite{peng2018sfv, xiong2021learning, das2020model}, image translation  and inpainting \cite{liu2018imitation,sharma2019third,smith2019avid, bahl2022human}, learning object centric representations \cite{pirk2019online, sermanet2018time}, simulators \cite{petrik2020learning,bonardi2020learning} and meta learning~\cite{yu2018one}.
Contrary to these prior works that often consider closely aligned human videos and robot environments e.g. humans demonstrating the task in the same lab environment as the robot, we assume that human video instructions are collected "in-the-wild" and therefore there exists a large domain gap between human videos and robot workspace.

Recent works  attempt to perform model pretraining from large-scale human videos in order to get good image representations for robotic control \cite{nair2022r3m, xiao2022masked}.
Other works have shown that we can infer states and actions from diverse videos and use it for reinforcement learning \cite{edwards2019perceptual, schmeckpeper2020reinforcement, schmeckpeper2020learning, seo2022reinforcement}.
In our work, we show that encoders trained on large datasets of videos for human action recognition provide good task embeddings for robotic manipulation.

Prior works have also considered leveraging large datasets of human videos to learn reward functions for robotics manipulations \cite{shao2021concept2robot, chen2021learning}.
The most relevant work to ours is Domain-agnostic Video Discriminator (DVD)~\cite{chen2021learning} which also tackles the problem of commanding robots using in-the-wild human videos.
DVD learns a video similarity by training a discriminator network to classify whether two videos are performing the same task on both annotated robot demonstrations and a subset of SSv2 videos.
The similarity score is then used as a reward for planning with an action-conditioned video generation method \cite{babaeizadeh2017stochastic} trained on randomly collected robot experience.
In contrast, our approach does not require any annotated robot demonstration and can accommodate both randomly collected robot experience and expert demonstrations.
\ec{Moreover,~\cite{chen2021learning} plans several sub-trajectories as high-dimensional synthetic videos per episode whereas our approach plans a full trajectory in the embedding space of robot videos.}

Contrastive learning on large-scale datasets has led to significant progresses in a range of computer vision tasks~\cite{chen2020simple, he2020momentum}.
Contrastive learning has also been used to learn language-conditioned policies by training on large scale datasets of images \cite{radford2021learning, shridhar2022cliport} and videos \cite{fan2022minedojo}.
In this work, we leverage Supervised Contrastive Learning \cite{khosla2020supervised} for human action recognition on the SSv2 dataset to learn a mapping between human video instructions and robotic manipulation.

Our work falls under outcome-conditioned action regression \cite{lynch2020learning, emmons2021rvs, furuta2021generalized}.
Such works cast reinforcement learning as learning policies conditioned on trajectory information such as future returns \cite{kumar2019reward, schmidhuber2019reinforcement,srivastava2019training}, many previous timesteps \cite{chen2021decision, janner2021offline} or a desired goal-configuration \cite{ding2019goal, ghosh2019learning, lynch2020learning, emmons2021rvs} to learn multi-task policies.
In contrast, our policy is conditioned on a video embedding of the full robot trajectory.
\cite{jang2022bc} also showed that we can learn a video-conditioned controllers by conditioning the policy on both a video of the full robot trajectory and paired sets of human videos appropriately collected for each task.
Other works \cite{lynch2020language, nair2022learning} show that pairing a small subset of robot data with language instructions enables generalizable language-conditioned task execution.
Our method does not require any robot data paired with human instructions and, hence, can learn generic policies from unlabelled robot datasets.
\begin{figure*}[t]
  \centering
  \vspace{.2cm}
  \includegraphics[width=1.0\linewidth]{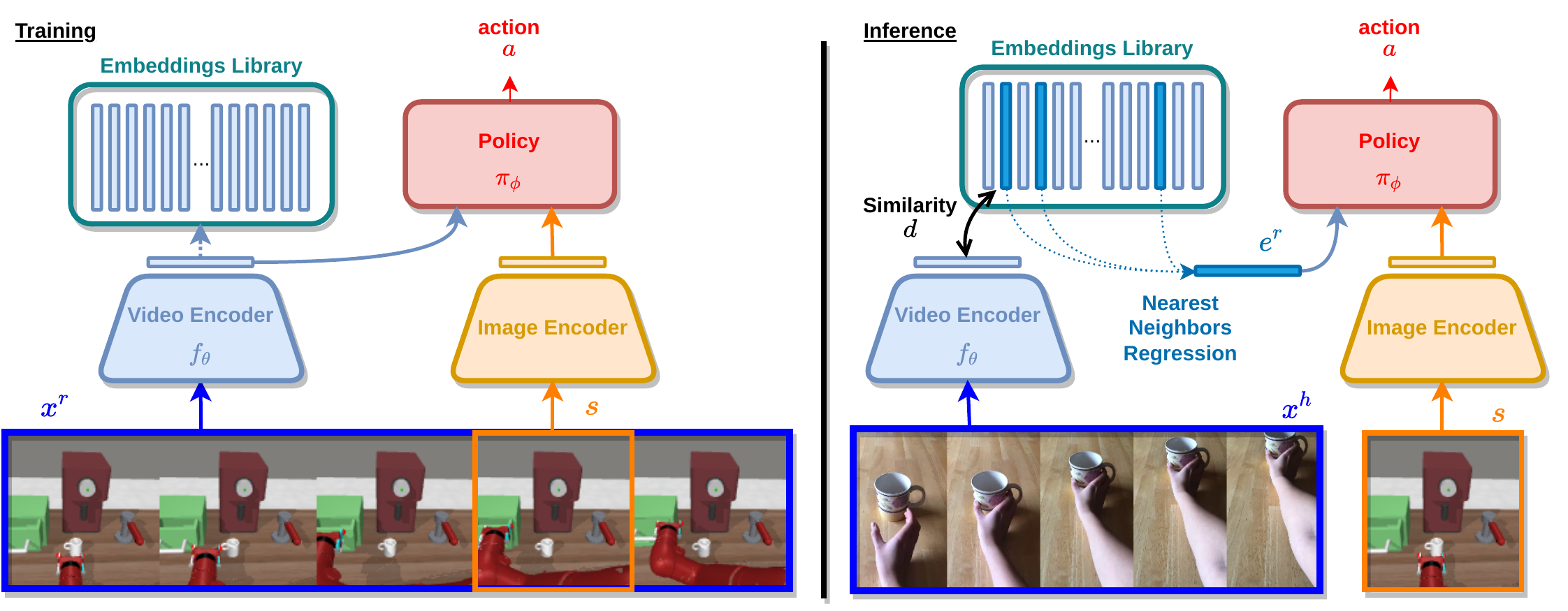}
  \vspace{-0.9cm} 
  \caption{
  (Left) During training, we learn a manipulation policy conditioned on robot video embeddings of the full robot trajectories from the robot dataset.
  At the same time, the robot video embedding of each trajectory in the robot dataset is added to an embeddings library.
  (Right) At inference, we encode the human video instruction into a human video embedding.
  We then average the robot embeddings from the library that have highest cosine similarity to the human embedding into a selected robot embedding.
  Finally, we execute the policy conditioned on this selected embedding.
  }
  \label{fig:PolicyLearning}
  \vspace{-0.5cm} 
\end{figure*}

\section{Method} \label{sec:method}

In Section~\ref{subsec:ProblemSetting} we first present an overview of our approach that enables robots to mimic new tasks demonstrated by people in natural human environments.
We then detail how we learn a generic video-conditioned policy from randomly-generated demonstrations
in Section~\ref{subsec:ViP}.
We further describe how we condition our policy on human videos in Section~\ref{subsec:Inference}.
Finally, we describe how we learn a similarity function for matching human videos to robot roll-outs without using paired human-robot data 
in Section~\ref{subsec:SimMetric}.

\subsection{Method overview} \label{subsec:ProblemSetting}

We aim to perform a robot manipulation task conditioned on a human video in a vision-based multi-task environment.
At test time our system receives an input video of a previously unseen task performed by a person, such as pushing a mug or closing a drawer, and controls a robot arm with the intent of performing a similar task in the robot environment.

More formally, we consider a set of Markov Decision Processes (MDP) $\mathcal{MDP}_i = (\mathcal{S}, \mathcal{A}, \mathcal{R}_i, p)_i$ sharing the same observation space $\mathcal{S}$, action space $\mathcal{A}$ and dynamics $p$ but with different reward functions $\mathcal{R}_i$ corresponding to different tasks we would like to solve.
We do not assume that the reward functions $\mathcal{R}_i$ are observed. 
Instead, $\mathcal{R}_i$ must be inferred through a human video of the task $x^h \in \mathcal{X}$.
Our goal is to learn a video-conditioned controller $\pi(.|s, x^h)$ that predicts actions $a \in \mathcal{A}$ given  the current states $s \in \mathcal{S}$ and human videos $x^h \in \mathcal{X}$ to maximize the reward functions associated with the input human video.

There may exist a large domain gap between videos of humans and robots performing similar tasks.
To bridge this gap, we leverage the large-scale video dataset Something-Something-v2 (SSv2)~\cite{goyal2017something} with labeld human actions $D^h = \{ x^h_i, y^h_i \}_i$ where labels $y^h_i$ for videos $x^h_i$ correspond to different manipulation actions such as Opening Something, Moving Something Away from the Camera, etc.
Following DVD~\cite{chen2021learning}, we train a similarity function $d(., .)$ that  assigns high values to pairs of videos representing the same task and low values to video pairs of different tasks.
Unlike DVD, however, we learn such similarity without any annotated robot videos.
Given a human video, we use the learned similarity as a reward $R(.) = d (. , x^h)$ for our controller.

To learn the policy, we assume access to a dataset of unlabelled robot demonstrations $D^r = \{ x^r_i, (s_i^t)_t, (a^t_i)_t \}_i$, where 
$(s_i^t)_t$ is a sequence of robot observations, $(a^t_i)_t$ is the corresponding sequence of executed robot actions and $x^r_i$ is a video of the demonstration. 
For instance, if the observation space $\mathcal{S}$ only contains images (without e.g. proprioceptive information), the videos can simply correspond to the whole sequence of states in the trajectory $x^r_i= (s_i^t)_t$.
This offline dataset can be collected in many different ways: by expert demonstrators, rollouts of other policies, through teleoperation or by random data generation.
Importantly, we do not assume access to any further information about these demonstrations. 

Figure \ref{fig:PolicyLearning} presents an overview of our approach:
we leverage a video encoder $f_\theta$ trained for human action recognition on SSv2 and a similarity metric between videos.
During training, we learn a behavior cloning policy $\pi_\phi$ conditioned on robot embeddings given by the video encoder while storing all embeddings of the robot training dataset in a library.
At inference, we first use this library and the similarity metric to predict a robot embedding relevant to the human video instruction, then rollout the policy in the environment.

\begin{figure*}[t]
  \centering
  \vspace{.2cm}
  \includegraphics[width=1.0\linewidth]{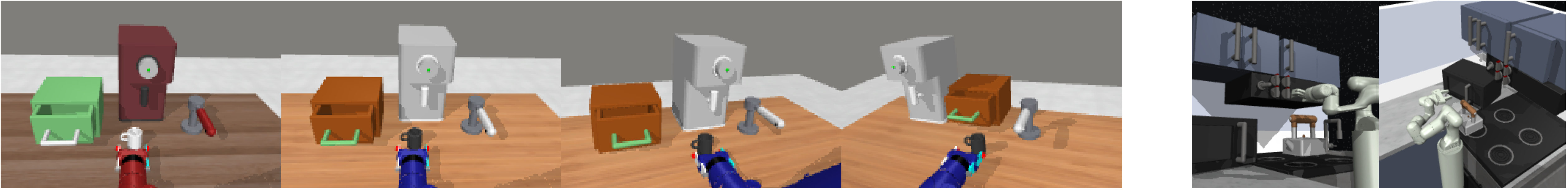}
  \vspace{-0.7cm} 
  \caption{
  Illustrations of environments used in our experiments. From left to right: TableTop env1-env4 followed by Kitchen left and right views.
  }
  \label{fig:Environments}
\vspace{-0.6cm} 
\end{figure*}

\subsection{Video-conditioned policy learning} \label{subsec:ViP}

We now describe how to encode a large array of behaviors into a single policy.
A key to our approach is the use of the embedding space of a video encoder pretrained on human videos to condition our policy.
This video embedding can be seen as a task embedding for our generic multi-task policy.
We will explain more in \ec{detail} how we can obtain meaningful video embeddings for control in \ec{Section} \ref{subsec:SimMetric}.

During training, we learn a policy $\pi_\phi$ to regress an action given the current state and a video embedding of a full robot trajectory.
Video embeddings act as context defining the global task. 
The policy is trained with behavior-cloning by minimizing the loss:
\begin{equation}
    \mathcal{L}_{\pi}(\phi) = - \mathbb{E}_{s, a, x^r \sim D^r} \log \pi_\phi (a |s, f_\theta(x^r)).
\end{equation}
At test time, this policy can be commanded to reproduce a robot video input $x^r$ by first encoding the video into an embedding $e^r$ and then executing the actions $a \sim \pi(.|s, e^r)$ predicted by the policy at each step $s$ visited during the roll-out. 
As we will see from experiments, however, using human videos to directly condition the policy results in poor performance due to the large domain gap between robot and human videos.
We therefore propose to first translate human videos to robot video embeddings as described in the next section. 

\subsection{Inference with human instructions} \label{subsec:Inference}
At inference, our first step is to translate the human video instruction $x^h$ into the robot video embedding $e^h$ that both (1)~corresponds to the target task and (2)~is in distribution for the robot policy.
While many methods can instantiate this, we choose to simply use nearest neighbors regression in the robot embedding space.
We first encode the videos contained in the robot dataset $D^r$ into a library of robot embeddings $D^e = (e^r_i = f_\theta(x^r_i))_i$.
We also encode the video instruction into a human embedding $e^h$.
We then perform $k$ nearest neighbors regression using the distance function $d$, by first computing all the distances between the human embedding and each embeddings in the library $d_i = d( e^h, e^r_i )$, then averaging the top $k$ embeddings of the library $e^r = \frac{1}{k} \sum_{\tilde{i} \in 1}^k e^r_{\tilde{i}}  $.
Finally, we perform policy rollout conditioned on this embedding.
\ec{As result, although the policy was trained with behavior cloning, this approach allows us to maximize the similarity of the robot trajectory to the video prompt at inference.}

\subsection{Learning a task similarity from human videos} \label{subsec:SimMetric}

Many choices of distance metric $d$ between videos can be used to match a human video to an appropriate robot embedding. 
In this work, we consider adapting Supervised Contrastive Learning~\cite{khosla2020supervised} to our video action recognition task on the Something-Something-v2 dataset.
We learn our video encoder $f_\theta(.) = f^{p}_\theta(f^{b}(.))$ composed of a backbone $f^{b}(.)$ which maps videos to representation vectors and a projection network $f^{p}_\theta(.)$ which maps representation vectors to embedding vectors $e$, such that embeddings from the same class are pulled closer together in cosine distance than embeddings from different classes.
As a result, the cosine distance characterizes the similarity between two videos.

In contrastive representation learning, the backbone and projection net are typically trained end-to-end from scratch and, after training, the projection net is discarded and a classifier head is learned on top of the representations.
Instead, we start from available pretrained backbones for SSv2 classification and simply train the projection net using the Supervised Contrastive Learning loss.
Given a batch of $N$ video/label pairs sampled from the SSv2 dataset $\{x^h_k, y^h_k\}_{k \in [1, .., N]} \sim D^h$, we build a multiview batch consisting of $2N$ pairs, $\{\tilde{x}^h_l, \tilde{y_l}\}_{l \in [1, .., 2N]}$,
where $\tilde{x}^h_{2k-1}$ and $\tilde{x}^h_{2k}$ are \ec{two random augmentations of video} $x^h_k$ and $\tilde{y}^h_{2k-1} =\tilde{y}^h_{2k} = y^h_k$.
We train $f_\theta$ to minimize the Supervised Contrastive Loss:
\begin{equation}\label{eq:supcon}
\begin{split}
    \mathcal{L}_{SupCon}(\theta) &= \\
    \sum_{i \in I} \frac{-1}{|P(i)|} \sum_{p \in P(i) } &\log \frac{\exp \left ( \langle f_\theta(\tilde{x}^h_i), f_\theta(\tilde{x}^h_p) \rangle / \tau \right )}{ \sum_{a \in A(i) } \exp \left ( \langle f_\theta(\tilde{x}^h_i) , f_\theta(\tilde{x}^h_a) \rangle / \tau \right ) }
\end{split}
\end{equation}
where $\langle ., . \rangle$ is the cosine similarity, $I = [1, .., 2N]$, $A(i) = I \backslash \{i\}$, $P(i) = \{ p \in A(i) : \tilde{y}^h_p = \tilde{y}^h_i \}$ and $\tau$ is a hyperparameter.
After training, we use the distance $d(x^h, x^r) = \langle f_\theta(x^h), f_\theta(x^r) \rangle$ between videos $x^h$ and $x^r$ as measure of similarity that focuses on the semantic aspects of the video.
As we show in the experimental section, despite being trained only on human videos, this similarity metric generalizes to robot manipulation tasks.
In our experiments, we use the same video backbone as \cite{shao2021concept2robot} and \cite{chen2021learning}.

Alternative options for $d$ include the DVD similarity network \cite{chen2021learning}, which learns to classify whether two videos correspond to the same task and uses the classification score between human and robot videos as a distance metric $d$ on top of the video encoder $f^b$.
While our approach implicitly corresponds to the same classification objective, supervised contrastive learning readily embeds our encoder with a similarity metric.

\section{Experiments} \label{sec:experiments}

This section presents experiments validating our proposed approach and its training procedure. 
We evaluate our method on several tasks in two challenging environments and compare results to the state-of-the-art method~\cite{chen2021learning}.
In particular, we present ablations of our method and show its advantage in zero-shot settings, i.e.~for tasks that have not been observed during training. 

\subsection{Experimental Setup}

\begin{table*}[t]
\vspace{.2cm}
\caption{Results for the TableTop environment using robot demonstrations. DVD results are obtained from~\cite{chen2021learning} except the ones marked with "*" for which we run the code provided by the authors.}
\label{table:tabletop_withrobot}
\vspace{-0.2cm} 
\centering
\begin{small}
\begin{tabular}{ l | l | l |l | l | l | l }
\toprule
Setting & Method & env 1 & env 2 & env 3 & env 4 & Avg \\ 
\midrule
seen robot demos
& DVD~\cite{chen2021learning} & 65.2 \tiny{(4.1)} & 62.3 \tiny{(2.8)} & 53.8 \tiny{(3.9)} & \textbf{39.6} \tiny{(0.7)} & 55.2 \tiny{(2.9)} \\
& ViP (DVD similarity) & \textbf{97.1} \tiny{(2.8)} & \textbf{72.0} \tiny{(15.9)} &  \textbf{82.1} \tiny{(16.4)} &  \textbf{42.0} \tiny{(11.2)} &  \textbf{73.3} \tiny{(11.6)} \\
\midrule
unseen robot demos
& DVD~\cite{chen2021learning} & 55.1 \tiny{(2.0)} & \textbf{51.1}* \tiny{(2.5)} & 38.4* \tiny{(1.6)} & 35.0* \tiny{(1.9)} & 44.9* \tiny{(2.0)} \\
& ViP (DVD similarity) & \textbf{68.2} \tiny{(11.2)} &  \textbf{55.2} \tiny{(13.0)} &  \textbf{60.3} \tiny{(8.0)} & \textbf{43.6} \tiny{(8.0)} & \textbf{56.8} \tiny{(10.0)} \\
\bottomrule
\end{tabular}
\end{small}
\vspace{-0.5cm} 
\end{table*}

For our experiments, we consider the TableTop environments introduced in~\cite{chen2021learning}.
The agent controls a robot arm in four variations of a simulated environment containing a drawer, a cup in front of a coffee machine and a faucet handle. 
The four environments differ by object positions, camera locations and colors.
Robot experience for policy training is collected by controlling the robot end effector to go through three keypoints randomly sampled in the environment.
We follow the experimental setup of~\cite{chen2021learning} and consider three tasks: close the drawer, push the cup and turn the faucet to the right. 
The image encoder representing the current state of the environment is learned end-to-end together with the policy.
These environments challenge the ability of our approach to generate meaningful manipulation skills from random robot trajectories. 

Furthermore, we consider the Kitchen environment initially introduced in~\cite{gupta2019relay}.
The agent controls a robot arm with a clamp in a kitchen environment containing diverse objects: a microwave, three doors, a kettle and light and knob switches.
We consider 3 opening tasks: open the microwave, open the left door and open the sliding door. 
Robot demonstrations accomplishing these tasks are available from~\cite{nair2022r3m}. 
However, these demonstrations are not annotated and the agent learns jointly from all these demonstrations without knowing what each demonstration accomplishes.
For the Kitchen environment we train the policy on top of R3M image representations following~\cite{nair2022r3m} using all demonstrations.
This environment is challenging as it requires to distinguish which task is the one intended by the user among very similar opening tasks.

Figure \ref{fig:Environments} illustrates the environments.
Following~\cite{chen2021learning}, for each of these tasks, we consider three different human video instructions collected in diverse environments to prompt our method at inference.
Note that, in both environments, our approach can use the same video encoder and similarity metric.

\subsection{Comparison to DVD~\cite{chen2021learning}}

We first validate our video-conditioned control policy and compare it to the planning method of DVD~\cite{chen2021learning} on the TableTop environments in the following settings:
\begin{itemize}
    \item \textit{Seen robot demos}: the video similarity is trained on a subset of task-related SSv2 classes and on labelled videos of robot demonstrations for the target tasks collected in env~1 and the rearranged version of env~1.
    \item \textit{Unseen robot demos}: the video similarity is trained on a subset of task-related SSv2 classes and on labelled videos of robot demonstrations corresponding to non-target tasks in env~1 and the rearranged version of env~1.
\end{itemize}
We compare our approach to~\cite{chen2021learning} that uses the DVD similarity on top of a learned action-conditioned video prediction model for control~\cite{babaeizadeh2017stochastic}.
For the \textit{Seen robot demos} setting we take results reported in~\cite{chen2021learning}.
For the \textit{Uneen robot demos}, we take the best success rate reported in~\cite{chen2021learning} for environment 1 and run the code of the authors to obtain results for environments 2, 3 and 4 \ec{(marked with *)}.
For \ec{a} fair comparison of our control policy, we here run ViP using the same video similarity function as in DVD \ec{which was trained on paired data between human videos and robot demonstrations}. 

Table \ref{table:tabletop_withrobot} presents results for the TableTop environments where, for each environment, we report the average success rates across the three tasks and three human video instructions.
ViP control policy outperforms DVD in both settings and \ec{most} environments.
\ec{We believe these improvements are due to comparing the human video prompt to full robot trajectories instead of shorter sub-trajectories and not relying on synthetic video generation for control as in~\cite{chen2021learning}.}
\ec{To demonstrate stability of our approach, we report results over four training seeds while~\cite{chen2021learning} reports results over evaluation runs.
}
Moreover, our method is significantly faster than DVD as it operates in the space of pre-computed embeddings, while DVD requires to execute its video prediction model for each sampled trajectory.
As result, in our experiments ViP requires less than 1 second to complete an episode in the TableTop environment while it takes more than 16 seconds to complete the same task with DVD.

We observe that the gains obtained by ViP compared to DVD are lower in the \textit{unseen robot demos} setting.
This can be attributed to the fact that our approach relies more on human videos of the target tasks to achieve good results.

\subsection{ViP without paired data}

While DVD uses paired data with robot and human videos demonstrating execution of similar tasks, obtaining such data is cumbersome at scale, see discussion in Section \ref{sec:introduction}.
In this section we demonstrate that ViP is able to cope with a more challenging setting where the training of video similarity is done using pairs of human videos only and without access to any robot data.

\begin{table*}
\vspace{.2cm}
\caption{Results for the TableTop environment without paired data. DVD results are obtained by running the code of the authors.}
\label{table:tabletop_human}
\vspace{-0.2cm} 
\centering
\begin{small}
\begin{tabular}{  l | l |l | l | l | l }
\toprule
Method & env 1 & env 2 & env 3 & env 4 & Avg \\ 
\midrule
Random & 25.6 & 25.1 & 22.2 & 18.2 & 22.8 \\
Human video as input & 22.7 \tiny{(7.9)} & 32.9 \tiny{(12.9)} & 11.2 \tiny{(7.8)} & 14.3 \tiny{(9.6)} & 20.3 \tiny{(9.6)} \\
DVD\cite{chen2021learning} & 43.0* \tiny{(2.6)} & 44.3* \tiny{(2.0)} & 32.8* \tiny{(1.7)} & 27.0* \tiny{(2.6)} & 36.8* \tiny{(2.2)} \\
ViP (Cosine distance in repr. space) & 36.9 \tiny{(4.5)} & 66.4 \tiny{(17.5)} & 39.3 \tiny{(15.0)} & 36.4 \tiny{(15.4)} & 44.8 \tiny{(13.1)} \\
ViP (DVD similarity) & 50.8 \tiny{(10.2)} & 67.3 \tiny{(17.4)} & \textbf{72.6} \tiny{(16.3)} & \textbf{53.4} \tiny{(9.6)} &  61.0 \tiny{(13.4)} \\
ViP (Ours) &  \textbf{79.9} \tiny{(11.4)} &  \textbf{77.9} \tiny{(16.2)} &  \textbf{70.6} \tiny{(17.8)} & \textbf{56.9} \tiny{(8.7)} & \textbf{71.3} \tiny{(13.5)} \\
\bottomrule
\end{tabular}
\end{small}
\vspace{-0.5cm} 
\end{table*}

Table \ref{table:tabletop_human} presents our results.
In this setting, \ec{our approach (\textit{ViP (Ours)})} significantly outperforms DVD, succeeding to perform intended tasks more than $70\%$ of the time on average across environments, while DVD performs only slightly better than chance.
Moreover, we can see that our results in this setting are similar to results in Table~\ref{table:tabletop_withrobot} where paired robot demonstrations are used for similarity learning. 
Indeed, having a global understanding of full robot trajectories makes our action recognition approach less reliant on robot demos, which are more important when planning for short horizon \ec{as in DVD.}
Finally, ViP with the DVD similarity metric also performs well, showing that our approach can accommodate alternative choices of video similarity. 
Qualitative results of ViP are presented in Figure~\ref{fig:Teaser} and in the supplementary video\ec{, where we show that failure cases include failing to manipulate the desired object and failure to perform the manipulation task.}

We further compare our approach to a baseline where we bypass our translation module and directly use human video embeddings to condition the ViP policy (\textit{Human video as input}).
Since ViP was only trained on robot video embeddings, embeddings of human videos are out of its training distribution.
As result, the policy fails to perform the intended tasks. This highlights the importance of our translation procedure which matches human videos to a library of robot demonstrations. 

\begin{figure}
  \centering
  \includegraphics[width=1.0\linewidth]{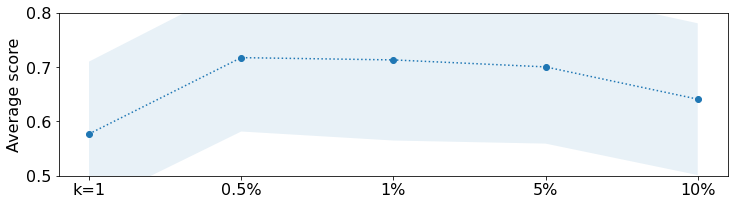}
  \vspace{-0.8cm} 
  \caption{
  Average success rate achieved by the policy on the TableTop environments for different values of $k$.
  We compare $k=1$ against values of $k$ corresponding to different percentages of the library size.
  }
  \label{fig:AblationKNN}
  \vspace{-0.6cm} 
\end{figure}

In Figure \ref{fig:AblationKNN} we evaluate the sensitivity of our method to the number of nearest neighbor samples $k$ used for translation of human videos in Section~\ref{subsec:Inference}.
High values of $k$ ensure that the generated embedding $e^r$ is an average of many appropriate robot embeddings, whereas low values of $k$ prioritize maximising the similarity score between the predicted robot embedding and the human instruction embedding.
While our approach is relatively robust to this hyperparameter, too high values of $k$ result in lower performance.
On the other side, for $k=1$, meaning that we choose a robot demonstration with the highest similarity to the human video, the performance drops.
We \ec{hypothesise} that performing a nearest-neighbor regression has a regularization effect, as naively maximizing the similarity might result in choosing an adversarial robot embedding that is misleadingly 
for the target task.
We set $k$ to $1\%$ of the library size for the TableTop environments and to $0.5\%$ for the Kitchen environment.

In Table \ref{table:tabletop_human} we also ablate an alternative choice for video similarity where video embeddings obtained from the SSv2 action classification network are directly compared using a cosine distance measure \textit{(Cosine distance in repr. space)}. 
Using such a naive video similarity together with ViP results in superior performance compared to DVD, however, it is outperformed by \ec{both the DVD similarity with ViP (\textit{ViP (DVD similarity)})  and our full approach, highlighting the importance of training an explicit distance between videos for the success of our method}.

\subsection{Kitchen environment}

We evaluate our approach on the kitchen environments in the zero-shot setting.
We compare our approach against a version of our method where we select a robot demonstration that solves the target task (\textit{ViP (Oracle)}) as well as a version that randomly selects robot demonstrations from the library (\textit{ViP (Random)}).
We also compare against \ec{single-task} R3M \cite{nair2022r3m}, where we trained specialized policies for each task using all the demonstrations available.

We report the success rates for each task averaged over the 3 corresponding human video instructions across 4 training seeds in Table \ref{table:kitchen_comparison}.
Comparison between R3M and \textit{Oracle} shows that our approach successfully learns from multi-task data for precise manipulation skills, achieving competitive results to R3M trained for single tasks.
\ec{When prompted with a human video instruction in the left camera view (L),} our approach successfully opens the correct door when presented with human videos of opening common doors and microwaves. However, the method fails to map human videos of opening sliding doors to appropriate robot skills.
On the right camera view (R), our approach struggles more, often opening the wrong object when prompted with videos of sliding doors and microwaves.
These results show the limit of training on SSv2, where different human videos of opening something are encouraged to be grouped together by our similarity metric disregarding manipulated objects.

\begin{table}[h]
\vspace{-0.2cm} 
\caption{Results on the Kitchen environment}
\label{table:kitchen_comparison}
\vspace{-0.2cm} 
\centering
\begin{small}
\begin{tabular}{ l | l | l |l | l }
\toprule
 & Method & left door & sliding door & microwave \\
\midrule
L
&ViP (Random) &  22.5 \tiny{(2.7)} & 29.8 \tiny{(0.4)} & 13.5 \tiny{(4.7)} \\
&ViP (Oracle) &83.2 \tiny{(16.5)} & 99.2 \tiny{(0.8)} & 43.0 \tiny{(10.2)} \\
&R3M\cite{nair2022r3m} & 63.8 \tiny{(10.3)} & 100 \tiny{(0.0)} & 53.8 \tiny{(11.0)} \\
\cmidrule{2-5}
&ViP (Ours) & 69.4 \tiny{(11.0)} & 41.7 \tiny{(27.6)} & 25.1 \tiny{(8.5)} \\ 
\midrule
R
&ViP (Random) & 22.0 \tiny{(2.9)} & 30.0 \tiny{(0.0)} & 8.5 \tiny{(3.0)} \\
&ViP (Oracle) & 92.0 \tiny{(7.4)} & 99.8 \tiny{(0.4)} & 27.2 \tiny{(9.1)} \\
&R3M\cite{nair2022r3m} & 57.5 \tiny{(2.9)} & 100 \tiny{(0.0)} & 30.0 \tiny{(0.1)} \\
\cmidrule{2-5}
&ViP (Ours) & 66.7 \tiny{(3.3)} & 41.7 \tiny{(14.4)} & 6.6 \tiny{(2.9)} \\ 
\bottomrule
\end{tabular}
\end{small}
\vspace{-0.4cm} 
\end{table}

\section{Conclusion}\label{sec:conclusion}

We propose ViP, a method that learns to map human video instructions to robot skills in a zero-shot manner.
We show that by conditioning on video embeddings, we can learn from multi-task robot data without supervision and prompt our policy with an unseen human video instruction at test time.
As a step further towards less supervision in data-driven robotics, we demonstrate that, by training on large datasets of diverse labelled human videos, we don't need to pair human instructions to robot data during training.

Our experiments demonstrate that ViP can accommodate many different similarity metrics between human instructions and robot manipulations.
Future work could explore other forms of similarities, such as mapping language instructions to robotic skills.



\small{
\textbf{Acknowledgements:} This work was funded in part by the French government under management of Agence Nationale de la Recherche as part of the ”Investissements d’avenir” program, reference ANR19-P3IA-0001 (PRAIRIE 3IA Institute), the ANR project VideoPredict, reference ANR-21-FAI1-0002-01, and 
the Louis Vuitton ENS Chair on Artificial Intelligence.
This work was granted access to the HPC resources of IDRIS under the allocation 2021-AD011012947 made by GENCI. 
We would like to thank Mathis Clautrier and Minttu Alakuijala for helpful feedbacks.

}

\bibliographystyle{IEEEtran}
\bibliography{main}



\end{document}